\pdfoutput=1
\documentclass[11pt]{article}

\usepackage[]{acl}

\usepackage{times}
\usepackage{latexsym}

\usepackage[T1]{fontenc}
\usepackage[utf8]{inputenc}

\usepackage{microtype}

\usepackage{inconsolata}

\usepackage{algorithm,algorithmic}
\usepackage{microtype}

\usepackage{amsmath}
\usepackage{cleveref}
\usepackage{xcolor, color}
\usepackage{graphicx}
\usepackage{epstopdf}
\usepackage{multirow}
\usepackage{booktabs}
\usepackage{comment}
\usepackage{float}
\usepackage{enumitem}
\usepackage{ctable} % for \specialrule command
\usepackage{makecell}
\usepackage{diagbox}

\title{Cobra Effect in Reference-Free Image Captioning Metrics}

\author{
    Zheng Ma \textsuperscript{\rm 1},
    Changxin Wang\textsuperscript{\rm 1},
    Yawen Ouyang\textsuperscript{\rm 2},
    Fei Zhao\textsuperscript{\rm 1,\rm 3}, \\
    \textbf{Jianbing Zhang\textsuperscript{\rm 1,\rm 3}}\thanks{Corresponding Author.}\textbf{,}
    \textbf{Shujian Huang}\textsuperscript{\rm 1}\textbf{,}
    \textbf{Jiajun Chen}\textsuperscript{\rm 1} \\
    \textsuperscript{\rm 1} National Key Laboratory for Novel Software Technology, Nanjing University, China\\
    \textsuperscript{\rm 2} Institute for AI Industry Research (AIR), Tsinghua University \\
    \textsuperscript{\rm 3} School of Artificial Intelligence, Nanjing University, China \\
    \texttt{\{maz, cx.wang, zhaof\}@smail.nju.edu.cn}, \quad
    \texttt{ouyangyawen@air.tinghua.edu.cn} \\
    \texttt{\{zjb, huangsj, chenjj\}@nju.edu.cn}
}
\begin{document}
\maketitle
\begin{abstract}
Evaluating the compatibility between textual descriptions and corresponding images represents a core endeavor within multi-modal research. 
In recent years, a proliferation of reference-free methods, leveraging visual-language pre-trained models (VLMs), has emerged. 
Empirical evidence has substantiated that these innovative approaches exhibit a higher correlation with human judgment, marking a significant advancement in the field.
However, does a higher correlation with human evaluations alone sufficiently denote the
complete of a metric?
In response to this question, in this paper, we study if there are any deficiencies in reference-free metrics.
Specifically, inspired by the Cobra Effect, we utilize metric scores as rewards to direct the captioning model toward generating descriptions that closely align with the metric's criteria. If a certain metric has flaws, it will be exploited by the model and reflected in the generated sentences.
Our findings reveal that descriptions guided by these metrics contain significant flaws, e.g. incoherent statements and excessive repetition.
Subsequently, we propose a novel method termed Self-Improving to rectify the identified shortcomings within these metrics. 
We employ GPT-4V as an evaluative tool to assess generated sentences and the result reveals that our approach achieves state-of-the-art (SOTA) performance.
In addition, we also introduce a challenging evaluation benchmark called Flaws Caption to evaluate reference-free image captioning metrics comprehensively.

\end{abstract}

\begin{figure}[h!]
    \setlength{\belowcaptionskip}{-0.5cm}
    \centering
    \includegraphics[width=0.48\textwidth]{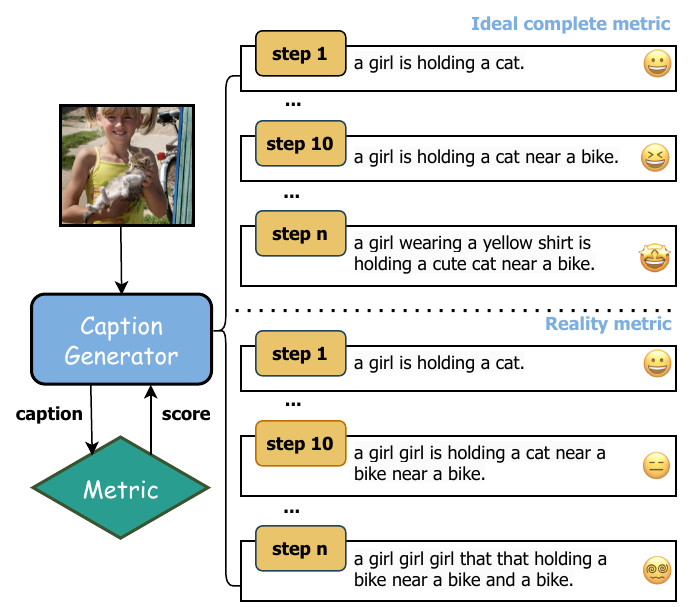}
    \caption{Top: Ideal state of the sentence generation process; Bottom: real situation of the sentence generation process.}
    \label{fig:intro}
\end{figure}

\section{Introduction}
Assessing the congruence between images and their descriptions is pivotal in numerous foundational multi-modal tasks, such as image captioning \citep{DBLP:conf/icml/WangYMLBLMZZY22, DBLP:conf/nlpcc/MaWHZZ23, DBLP:conf/mm/ChengSMZZZ23} and image-text retrieval \citep{bottom-up, DBLP:conf/ijcai/ShiJLND19, DBLP:conf/iccv/LiZLLF19}.
The burgeoning development of visual-language pre-trained models (VLMs) \cite{DBLP:conf/icml/RadfordKHRGASAM21, DBLP:conf/nips/LuBPL19, DBLP:conf/icml/0001LXH22, DBLP:conf/icml/WangYMLBLMZZY22} alongside large-scale visual-language models (LVLMs) \cite{DBLP:journals/corr/abs-2304-08485, DBLP:journals/corr/abs-2308-12966, DBLP:journals/corr/abs-2311-12793} has significantly propelled the precise evaluation of image-description compatibility. 
Traditionally, the assessment methodology relied heavily on manual reference creation, followed by a comparison of these references with candidate descriptions using metrics such as BLEU \cite{DBLP:conf/acl/PapineniRWZ02} and CIDEr \cite{DBLP:conf/cvpr/VedantamZP15}, and most prior works use these metrics as comparison criteria.
However, the manual approach often fails to encapsulate the entirety of information depicted in images and leads to the issue of excessive penalization \citep{DBLP:conf/emnlp/JiangHZWZGDG19}.
With the advent of advanced multi-modal matching technologies, a direct assessment of text quality based solely on the corresponding images has become viable. 
This evolution has given rise to several exemplary reference-free image captioning metrics \citep{DBLP:conf/acl/Lee0DBJ20, DBLP:journals/corr/abs-2104-08718} based on VLMs, which have demonstrated substantial correlation with human judgment. Consequently, these reference-free metrics, characterized by their streamlined approach and superior efficacy, have heralded a new paradigm within the domain of image captioning metrics.

A question arises: does a higher correlation with human evaluations alone sufficiently denote the comprehensiveness of a metric? 
In response to this question, our study leverages reference-free image captioning metrics as rewards within a reinforcement learning framework to steer the model to generate descriptions that more closely adhere to these metrics. 
Our idea is inspired by the Cobra Effect \footnote{The concept of the cobra effect originated from an anecdote in India under British governance, attributed to economist Horst Siebert. Alarmed by the high population of venomous cobras in Delhi, the British authorities introduced a reward system for each cobra killed. This plan initially achieved its intended effect, with many snakes being exterminated for the bounty. Over time, however, individuals started breeding cobras as a source of income, undermining the strategy's initial success. (Refer to \url{https://en.wikipedia.org/wiki/Perverse_incentive})} \citep{varpio2017shedding},
if metrics are used as rewards and if there are unreasonable issues within these metrics, then these problems will be exposed similarly to the Cobra Effect. 
Specifically, a complete metric ought to direct the model towards generating superior descriptions. This entails the metric's continuous evaluation of generated descriptions, distinguishing between those that meet the established criteria and those that do not. 
The process ensures an incremental alignment of generated descriptions with the metric's standards. As depicted in ~\Cref{fig:intro} at the top, with the sustained guidance of the metric, there is a notable improvement in the scores attributed to the model-generated descriptions. Simultaneously, the content of these descriptions becomes increasingly accurate and encompassing.

We conduct experiments utilizing current popular metrics to ascertain their efficacy in enhancing the metric score of model-generated descriptions. 
Our results indicate that, following reinforcement learning (RL), there is a noticeable increase in the scores of the generated descriptions, which is consistent with our expectations.
Surprisingly, a detailed examination of the specific sentences revealed that they became largely unreadable, plagued by issues such as incoherence and disorganization, as depicted in ~\Cref{fig:intro} at the Bottom. 
After closely examining the results from the test set, we find that nearly all generated sentences exhibit these issues (We provide more examples in \Cref{tb:gen_sent}). 
\textbf{This observation suggests that these metrics fail to handle these issues, leading to generated sentences with these flaws.}

To address these issues, we introduce a novel approach termed Self-Improving aimed at repairing metric deficiencies. 
The Self-Improving utilizes the sentences generated during the RL stage as negative examples to re-train the metric, which is a process of self-discovery of problems and self-repair.
After the Self-Improving phase, our repaired metric is also applied as a reward to guide the captioning model.
We utilize GPT-4V as an evaluative tool to score sentences generated by eight different models. The experiment suggests that sentences generated by our method, which incorporated the revised metrics as rewards, demonstrated superior quality.

In addition, to underscore the limitations inherent in existing mainstream reference-free metrics, we introduce a challenging benchmark named Flaws Caption, which requires models to discern the correct description from among a set of candidate sentences with interference. Experiments on Flaws Caption demonstrate that mainstream metrics are all below 60 points, but our Self-Improving method improves by 38.2 points over the current state-of-the-art method.

Summarizing our contributions, we present three key points:
\begin{itemize}
    \item{By studying the mainstream reference-free metrics, we disclose that these metrics enhance correlation with human judgment, but they universally manifest considerable deficiencies.}
    \item {We introduce a novel method called Self-Improving, wherein sentences generated through reinforcement learning serve as negative samples to address issues in the metrics. By re-training the metric and employing it to direct the generation of model-produced sentences, our approach secures the highest GPT-4V scores among compared metrics.}
    \item {We present a challenging evaluation benchmark, consisting of sentences with various interferences.}
\end{itemize}

\section{Related Work}
\subsection{Automatic Image captioning Metrics}
The automatic metrics for assessing how well a description corresponds to the visual content are essential in most multi-modal tasks.
Initially, metrics borrowed from machine translation and text summarization fields are adapted for image captioning, such as BLEU \citep{DBLP:conf/acl/PapineniRWZ02}, METEOR \citep{DBLP:conf/acl/BanerjeeL05} and ROUGE \citep{lin2004rouge}. 
subsequently, the specific metrics for image captioning metrics are proposed. \citet{DBLP:conf/cvpr/VedantamZP15} use Term Frequency-Inverse Document Frequency (TF-IDF) weighting to prioritize distinctive words in captions. \citet{DBLP:conf/eccv/AndersonFJG16} introduce a sophisticated approach by scene graph representations to evaluate the fidelity of captions to the content of images.
Nevertheless, the aforementioned methodologies predominantly hinge on n-gram analysis and depend heavily on Ground Truth comparisons, which leads to issues of over-penalty. Consequently, this recognition has spurred researchers to investigate metrics in a reference-free approach.

The advent of VLMs has catalyzed the proposal of numerous reference-free image captioning metrics predicated on these models, demonstrating a remarkable congruence with human evaluative judgments \citep{DBLP:conf/acl/Lee0DBJ20, DBLP:journals/corr/abs-2104-08718, DBLP:conf/acl/HuCZJ23}.
These methods enable the direct comparison of the similarity between images and candidate descriptions, obviating the need for reference assistance.
Although these methods exhibit a high correlation with human evaluations and offer considerable convenience, they overlook whether there are unexpected issues within these metrics.

\subsection{Flaws in Image-Text Matching}
\label{sec:flaws_of_vlm}
Recently, Image-text matching have been a pre-training task in VLMs and VLMs have achieved satisfactory performance in correseponding tasks\citep{DBLP:conf/nips/LuBPL19, DBLP:conf/icml/RadfordKHRGASAM21, DBLP:conf/icml/0001LXH22}, e.g. image-text retrieval. 
However, the matching capability of VLMs is called into question.
\citet{yuksekgonul2022and} discover that VLMs are prone to mistakes in associating objects with their attributes, and display a significant deficiency in sensitivity to sequence.
\citet{ma-etal-2022-probing} find that VLMs rely on objects in images and nouns in descriptions, and prefer specific sentence patterns and sentences with more nouns.
Some efforts have also been made to enhance the matching capabilities of VLMs.
\citet{DBLP:conf/acl/Lee0DBJ20} employ manually constructed negative samples to improve VLMs in image-text matching.
\citet{DBLP:conf/naacl/00010KDBB22} suggest that certain VLMs are not trained with MLM task \citep{DBLP:conf/naacl/DevlinCLT19} during pre-training, therefore utilizing manually constructed negative samples to enhance the grammatical abilities of VLMs.
Contrary to existing approaches, our research is primarily dedicated to pinpointing the deficiencies inherent in reference-free metrics. Furthermore, our proposed method is based on repairing the problems exposed by the metrics themselves.

\begin{figure*}[h!]
    \setlength{\belowcaptionskip}{-0.5cm}
    \centering
    \includegraphics[width=0.98\textwidth]{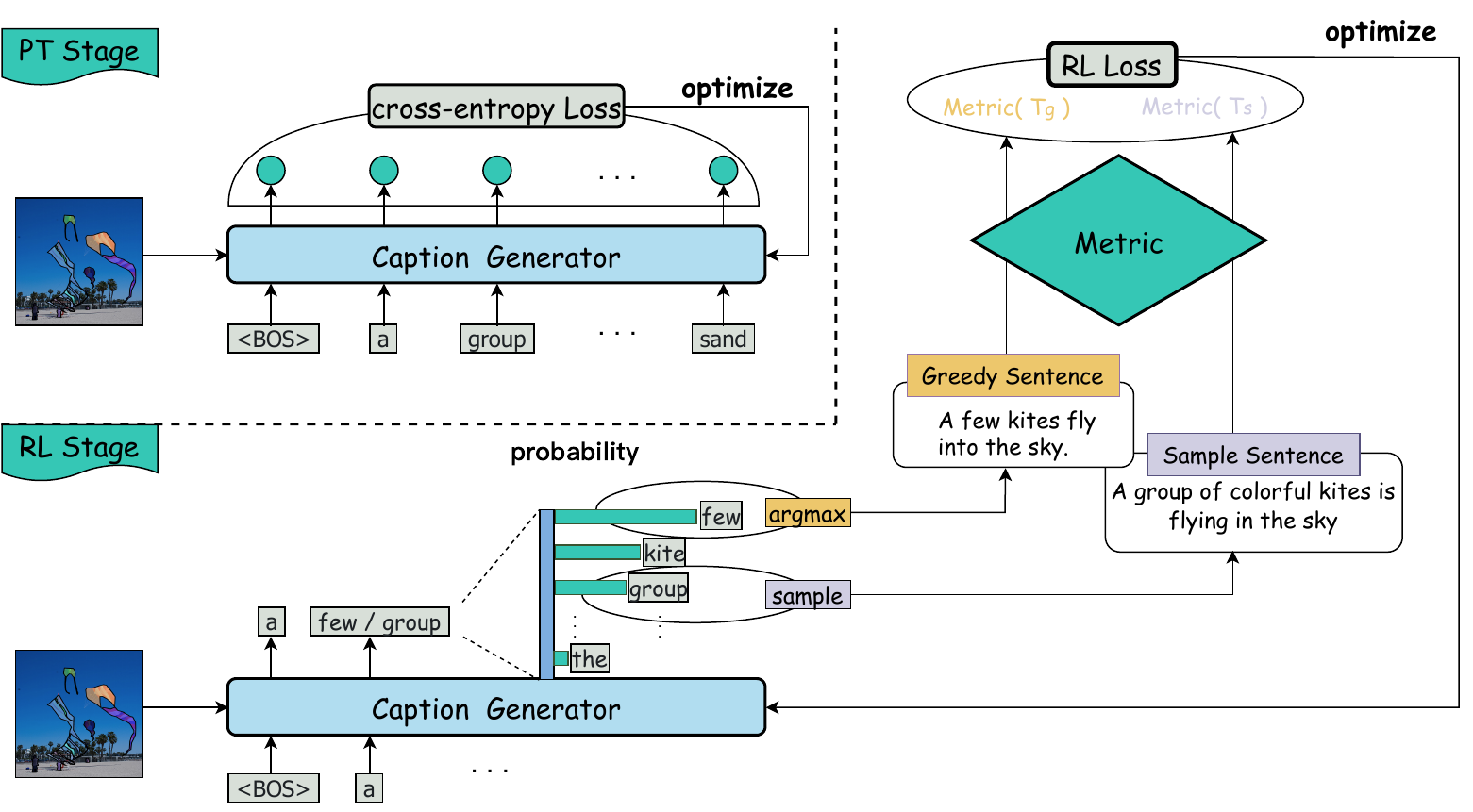}
    \caption{Flowchart of evaluation Reference-free metrics. PT Stage: the model is optimized using cross-entropy loss. RL Stage: we employ two distinct decoding strategies: greedy decoding, resulting in T$_{g}$, and sampling decoding, producing T$_{s}$. The objective is to evaluate the difference between these two generated sentences in terms of the specified metric.}
    \label{fig:rl_pt}
\end{figure*}

\section{Regarding Metrics as Rewards}
\label{sec:reward_method}
\subsection{Evaluated Reference-Free Metrics}
To comprehensively evaluate reference-free captioning metrics, we selected mainstream metrics, including UMIC, CLIPScore, BLIPScore, and InfoMetIC.

\noindent \textbf{UMIC} \citep{DBLP:conf/acl/Lee0DBJ20} is the first reference-free image captioning metric, which finetunes UNITER \citep{DBLP:conf/eccv/ChenLYK0G0020} via contrastive learning to distinguish Ground Truth and synthetic negative descriptions, So it can assess images and descriptions directly by the image-text matching scores.

\noindent \textbf{CLIPScore} \citep{DBLP:journals/corr/abs-2104-08718} directly utilize CLIP \citep{DBLP:conf/icml/RadfordKHRGASAM21} to extract visual and textual features and calculate the similarity between the vectors to measure the matching degree. 

\noindent \textbf{BLIPScore} is used as a comparative baseline in \citep{DBLP:journals/corr/abs-2305-11116}. Its algorithm is identical to CLIPScore, with the only difference being the substitution of BLIP \citep{DBLP:conf/icml/0001LXH22} for CLIP.

\noindent \textbf{InfoMetIC} \citep{DBLP:conf/acl/HuCZJ23} takes into account more fine-grained information, such as incorrect words and unmentioned image regions. we use InfoMetIC and InfoMetIC$_\text{plus}$ as rewards in our experiments.

\subsection{Identify Issues through Metrics Reward Feedback}
Our objective is to determine whether are there some unknown deficiencies in metrics that have a high similarity to human judgment. 
To thoroughly investigate this issue, in our approach, these metrics are utilized as rewards, and we adopt the Self-Critical Sequence Training (SCST) strategy \citep{DBLP:conf/cvpr/RennieMMRG17} to train the captioning model. 
This SCST process encompasses both the Pre-Training (PT) and Reinforcement Learning (RL) stages, as depicted in \Cref{fig:rl_pt}.

\paragraph{Pre-Training (PT) Stage.} 
Given the expansive action space for sentence generation, which aligns with the vocabulary size, the model's sampling efficiency is markedly limited without pre-training.
This limitation will hamper the model's ability to generate coherent sentences during the RL training stage.
To mitigate the above issue, we initially pre-train the captioning model using supervised learning and use cross-entropy between generated sentences and Ground Truth as the loss function. 
Let $I$ denote an image and $T = ({ w_1, w_2, \dots, w_l })$ represent a description, where $l$ is the total length of the description. The loss function for the pre-training stage can thus be expressed as:
\begin{align}
    \mathcal{L}_{pt} = -\sum_{t=1}^{l} \log p_{\theta}(w_{t}|w_{<t}, I) . 
\end{align}

% \begin{table}
% \centering
% \small 
% \resizebox{0.4\textwidth}{!}{
% \begin{tabular}{lcc}
% \hline
% \multirow{2}{*}{\textbf{Model}} & \multicolumn{2}{c}{\textbf{RL stage}} \\
%  & \textbf{Before } & \textbf{After}\\
% \hline

% UMIC & 66.5 & \textbf{74.0}   \\
% CLIPscore & 73.5 & \textbf{77.2} \\
% BLIPScore & 45.1 & \textbf{47.0} \\
% InfoMetIC & 40.3 & \textbf{46.3} \\
% InfoMetIC$_\text{plus}$  & 10.9 & \textbf{15.2} \\

% \hline

% \end{tabular}
% }
% \caption{Comparison before and after the RL stage across different metrics. After the RL stage, the scores of all metrics improved.}
% \label{tb:rl_results}
% \end{table}

\begin{table*}
\centering
\small 
\resizebox{0.8\textwidth}{!}{
\begin{tabular}{lccccc}
\toprule
\textbf{Stage} & \textbf{UMIC} & \textbf{CLIPscore}  & \textbf{BLIPScore}  & \textbf{InfoMetIC}  & \textbf{InfoMetIC$_\text{plus}$} \\ 
\midrule

Before & 66.5 & 73.5 & 45.1 & 40.3 & 10.9 \\
After & \textbf{74.0} & \textbf{77.2} & \textbf{47.0} & \textbf{46.3} & \textbf{15.2} \\

% UMIC & 66.5 & \textbf{74.0}   \\
% CLIPscore & 73.5 & \textbf{77.2} \\
% BLIPScore & 45.1 & \textbf{47.0} \\
% InfoMetIC & 40.3 & \textbf{46.3} \\
% InfoMetIC$_\text{plus}$  & 10.9 & \textbf{15.2} \\

\bottomrule

\end{tabular}
}
\caption{Comparison before and after the RL stage across different metrics. After the RL stage, the scores of all metrics improved.}
\label{tb:rl_results}
\end{table*}
\begin{table*}[h!]
\centering
\small
\resizebox{0.7\textwidth}{!}{
\begin{tabular}{lccccc}
\toprule
\textbf{Method} & \textbf{Bleu1} & \textbf{Bleu4} & \textbf{METEOR} & \textbf{ROUGE} & \textbf{CIDEr} \\
\midrule

% \midrule
UMIC & 49.5 & 14.8 & 24.8 & 43.0 & 35.0 \\
CLIPscore & 55.6 & 21.3 & 27.3 & 48.1 & 52.6 \\
BLIPScore & 58.3 & 20.8 & 26.9 & 49.3 & 68.4  \\
InfoMetIC & 53.6 & 19.7 & 26.9 & 47.0 & 47.5 \\
InfoMetIC$_\text{plus}$  & 51.7 & 18.8 & 26.9 & 45.9 & 38.2  \\\midrule
Before & \textbf{75.0} & \textbf{33.1} & \textbf{27.4} & \textbf{55.7} & \textbf{110.5}  \\
% & 20.8  & 22.4& 22.0 & 21.8  & 22.3 & 20.6
% \midrule
% Ground Truth (test) & 92.7 & 96.7 & 85.2 & 30.3 & 94.7 \\
\bottomrule
\end{tabular}
}
% \caption{Results on scores of reference-free and common-used image captioning metrics. Baseline denotes the vanilla Transformer model.}
\caption{Results on reference-based metrics for the methods before and after RL optimization using reference-free metrics.}

\label{tb:reference-based_results}
\end{table*}

\paragraph{Reinforcement Learning (RL) Stage.}
After the PT phase, the cross-entropy loss function is replaced with a metric-based reward system to further refine the model through reinforcement learning. 
In this process, the model first generates a sentence $T_g$ using a greedy algorithm, serving as the baseline for comparison:
\begin{align}
    w^{*}_{t} = argmax_{w_{t}} \ p_{\theta}(w_{t}|w^{*}_{<t}, I), 
\end{align}
where $w^{*}_{t}$ represents the word is selected at $t$ step by greedy algorithm. 

Subsequently, the model generates an alternative sentence through multinomial sampling, which is referred to as the sample sentence $T_s$:
\begin{align}
    w^{'}_{t} = sample \ p_{\theta}(w_{t}|w^{'}_{<t}, I), 
\end{align}
where $w^{'}_{t}$ represents the word is selected at $t$ step by multinomial sampling algorithm. 

The evaluation metric is then applied to compare the quality of the sample sentence against the greedy sentence:
\begin{align}
    \text{reward}_{diff} = \frac{metric(T_{s}) - metric(T_{g})}{|metric(T_{g})|},
\end{align}
where $T_s$, $T_g$ denote the sample sentence and greedy sentence respectively.

Should the metric score of the sample sentence exceed that of the baseline, the selection probability of the sample sentence for subsequent iterations is augmented. Conversely, a lower score results in a reduction of this probability. This procedural dynamic can be mathematically represented as follows:

\begin{align}
    \mathcal{L}_{rl}= - \text{reward}_{diff} \bigtriangledown \log \ p(T_{s}).
\end{align}
\begin{table*}[h!]
\centering
\resizebox{0.99\textwidth}{!}{
\scriptsize
\begin{tabular}{c|m{.7\textwidth}}
\toprule
\textbf{Image} & \textbf{Generated Captions} \\
\midrule
% Example 1

\multirow{7}{*}{\begin{minipage}{.16\textwidth}\includegraphics[width=\linewidth]{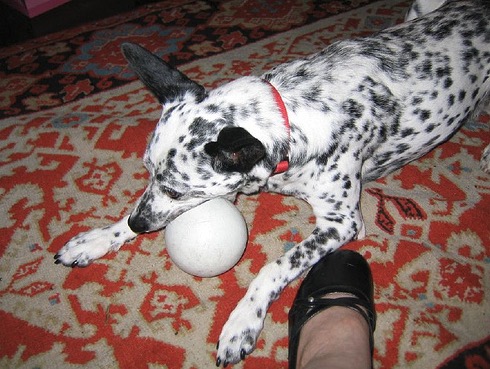}\end{minipage}} & \underline{\textbf{Baseline}} a dog laying on a rug with a tennis ball \\\cmidrule{2-2}
& \underline{\textbf{UMIC}} a black and white dog laying on a red and white blanket with a white ball with a white ball \\
& \underline{\textbf{CLIPScore}} a dog laying on a rug with a white ball on a rug with a white ball and white ball \\
& \underline{\textbf{BLIPScore}} a dog laying on a rug with a white ball with a white ball with a white \\
& \underline{\textbf{InfoMetIC}} a dog laying on a blanket with a white ball on a rug with a white ball \\
& \underline{\textbf{InfoMetIC$_\text{plus}$}} a black and white dog laying on a red and white blanket with a white ball on\\
\midrule

% Example 2
\multirow{8}{*}{\begin{minipage}{.16\textwidth}\includegraphics[width=\linewidth]{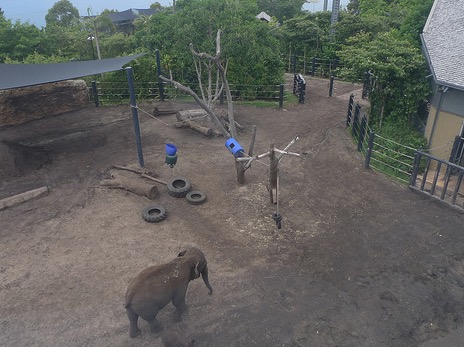}\end{minipage}} & \underline{\textbf{Baseline}} a baby elephant walking in a dirt area next to a barn \\\cmidrule{2-2}
& \underline{\textbf{UMIC}} a black and white dog laying on a red and white blanket with a white ball with a white ball \\
& \underline{\textbf{CLIPScore}} a small baby elephant walking in dirt near some buildings and a blue water can with some blue water can \\
& \underline{\textbf{BLIPScore}} a baby elephant walking next a baby elephant in a dirt area next a house and a house \\
& \underline{\textbf{InfoMetIC}} a small baby elephant walking next to a small baby elephant in a dirt area next \\
& \underline{\textbf{InfoMetIC$_\text{plus}$}} a small baby elephant walking next to a small baby elephant in a dirt area near a dirt area near\\
\midrule

% Example 3
\multirow{7}{*}{\begin{minipage}{.16\textwidth}\includegraphics[width=\linewidth]{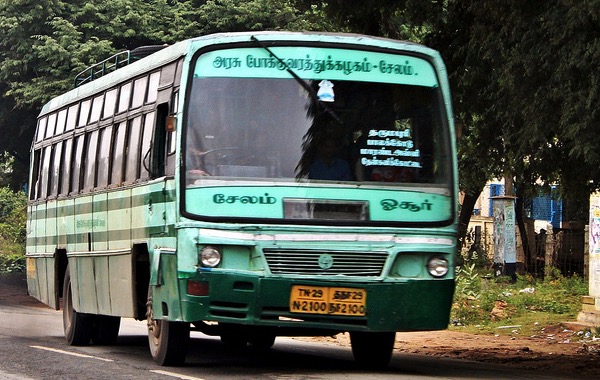}\end{minipage}} & \underline{\textbf{Baseline}} a green bus driving down a street next to a forest \\\cmidrule{2-2}
& \underline{\textbf{UMIC}} a green bus that is on the road near some trees and some trees and some green trees and some \\
& \underline{\textbf{CLIPScore}} a green bus parked on the side of a road next to a green tree with \\
& \underline{\textbf{BLIPScore}} a green bus that is parked on the street next a tree and a tree tree next the street next \\
& \underline{\textbf{InfoMetIC}} a green bus parked next to a tree on the side of a road next to a tree \\
& \underline{\textbf{InfoMetIC$_\text{plus}$}} a green bus parked on the side of a road next to a tree next to a tree on a\\

\bottomrule
\end{tabular}
}
\caption{The cases of sentences generated by captioning models, each guided by a distinct reference-free metric. Baseline refers to the implementation of a Vanilla Transformer model, serving as the standard for comparison.}
\label{tb:gen_sent}
\end{table*}

\subsection{Experimental Setup}
Our experiments are conducted on the MSCOCO dataset \citep{chen2015microsoft}. 
To sufficiently utilize MSCOCO, we adopt Karpathy's split \citep{karpathy2015deep}, re-allocating the dataset into subsets consisting of 113,287 images for training, 5,000 for validation, and 5,000 for testing. 
In our preprocessing stage, we analyze the frequency of word occurrences within the captions, excluding words that appear fewer than five times. 
This approach enabled us to construct a vocabulary comprising 9,487 words.
We employ a Vanilla Transformer \cite{transformer} as the backbone for our captioning model. To preprocess the images, we extract regions using a Faster R-CNN model \cite{bottom-up}, which is pre-trained on the Visual Genome dataset \cite{vg}. During the PT stage, the captioning model undergoes training for 15 epochs. Subsequently, in the RL stage, we extend the training for an additional 20 epochs, utilizing reference-free metric scores as rewards.

\subsection{Main Results}
\label{sec:main_results}
We provide the results of training with metric scores as rewards in \Cref{tb:rl_results}. It illustrates an enhancement in all metric scores following the reinforcement learning (RL) stage, suggesting that these metrics effectively direct the model toward generating descriptions more closely aligned with each respective metric's standards. 
However, there is a marked degradation in model performance across reference-based metrics\footnote{We compute these metrics by coco-caption (\url{https://github.com/tylin/coco-caption})} after the RL phase (shown in \Cref{tb:reference-based_results}). This downturn highlights a significant divergence between the model-generated sentences and the Ground Truth. To clarify this situation, we examine the specific generated sentences for all metrics. 
Surprisingly, we observe that almost generated sentences guided by metrics become strange, such as having many repeated words or phrases, as detailed in \Cref{tb:gen_sent}.
The results substantiate that these metrics have vulnerabilities in judging these issues, which leads to models trained with them as rewards generating sentences in this manner.
In \Cref{sec:fix_flaws}, we will introduce a novel strategy to repair these issues.

\section{Resolve Flaws by Self-Improving}
\label{sec:fix_flaws}
As highlighted in \Cref{sec:main_results}, employing mainstream metrics to direct the captioning model results in improved scores on these metrics, but the generated sentences have many issues. These issues underscore the underlying deficiencies of the metrics themselves. 
In this section, we propose a novel approach, termed Self-Improving, aimed at rectifying these metric-related deficiencies. This method leverages the model guided by the flawed metrics to generate negative samples. By identifying these problematic sentences, the model applies contrastive learning techniques to retrain and refine the metrics, thereby enhancing their reliability.

\subsection{Data Collection}
For our experiments, we utilized the MSCOCO dataset, for each image, one caption from the Ground Truth is randomly selected as a positive sample, while the sentence generated by the captioning model for the same image served as a negative sample. 
This procedure was applied across a total of 113,287 images, yielding 226,574 descriptions, each comprising one positive and one negative description. For evaluation, a subset of 5,000 images from the test set was employed.

To validate the efficacy of our proposed method, we adopt the widely used CLIPScore, a widely acknowledged method in the reference-free image captioning metric. We further enhanced the CLIP model by fine-tuning it with the positive samples from our specially constructed dataset, resulting in a variant we have denoted as CLIPScore$_{\text{ft}}$.

\begin{figure}[h!]
    \centering
    \includegraphics[width=0.49\textwidth]{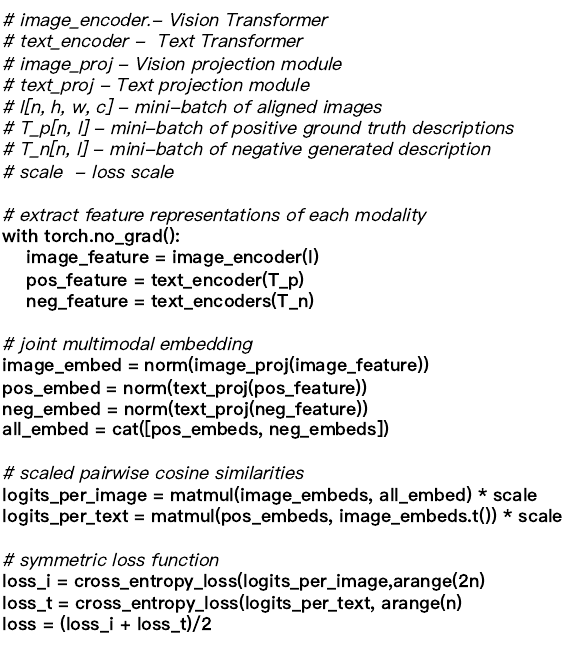}
    \caption{Pytorch-like pseudocode for the core of an implementation of Self-Improving based on CLIP. }
    \label{fig:pseudocode}
\end{figure}

\begin{table}
\centering
\small 
\resizebox{0.45\textwidth}{!}{
\begin{tabular}{lc}
\hline
\textbf{Metric} & \textbf{Kendall}\\
\hline
CLIPScore & 53.5 \\ 
CLIPScore$_\text{ft}$ & 57.4 \\
CLIPScore$_\text{self-imp}$ (ours) &  56.3 \\ 

\hline
\end{tabular}
}
\caption{ Kendall Correlation between human judgments and metrics on Flickr8K-Expert dataset.}
\label{tb:flickr8k_result}
\end{table}

\subsection{Self-Improving Method}
Our investigation has revealed that the sentences generated by the model illuminate the shortcomings of the current metrics. Consequently, a pressing question arises: how can we rectify these deficiencies within the model? An intuitive solution is to sensitize the model to the issues present in these sentences and subsequently recalibrate the metrics to mitigate these flaws.

In response, we introduce a pioneering strategy named Self-Improving. This approach capitalizes on the model, directed by the existing metrics, to generate negative samples, thereby enabling the model to identify and comprehend the inherent problems within these sentences. Subsequently, it applies contrastive learning techniques to re-train the model, aiming to overcome its original limitations. The procedural steps of the Self-Improving method are detailed in the pseudocode presented in \Cref{fig:pseudocode}.

After re-training, We use Flickr8K-Expert benchmark \citep{DBLP:conf/ijcai/HodoshYH15} to validate our model (CLIPScore${_\text{self-imp}}$) ability in mathcing. We report the results in \Cref{tb:flickr8k_result}. We observe that utilizing the MSCOCO dataset can improve the matching ability compared to vanilla CLIPScore. In addtion, we find CLIPScore$_{\text{ft}}$ performs slightly better than CLIPScore${_\text{self-imp}}$. This is because the Flickr8K-Expert benchmark does not include the flaws we find in \Cref{sec:main_results}. So we introduce a new challenging benchmark named Flaws Caption to assess reference-free image captioning metrics. We will provide a detailed description in \Cref{sec:flaws_caption}.

\begin{figure}[h!]
    \setlength{\belowcaptionskip}{-0.5cm}
    \centering
    \includegraphics[width=0.49\textwidth]{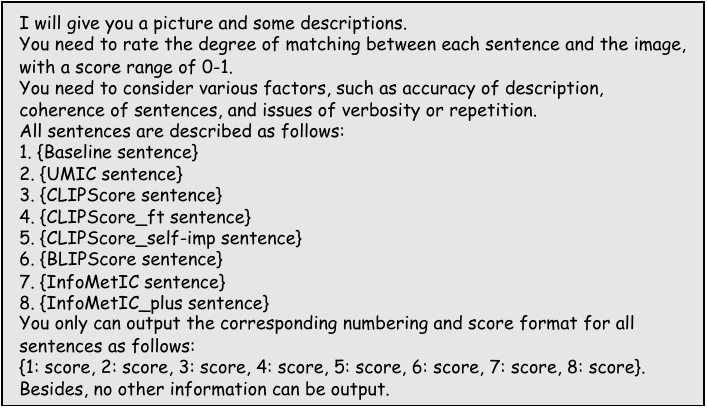}
    \caption{The prompt of the GPT-4V evaluator.}
    \label{fig:prompt}
\end{figure}

\begin{figure*}[h!]
    \centering
    \includegraphics[width=0.98\textwidth]{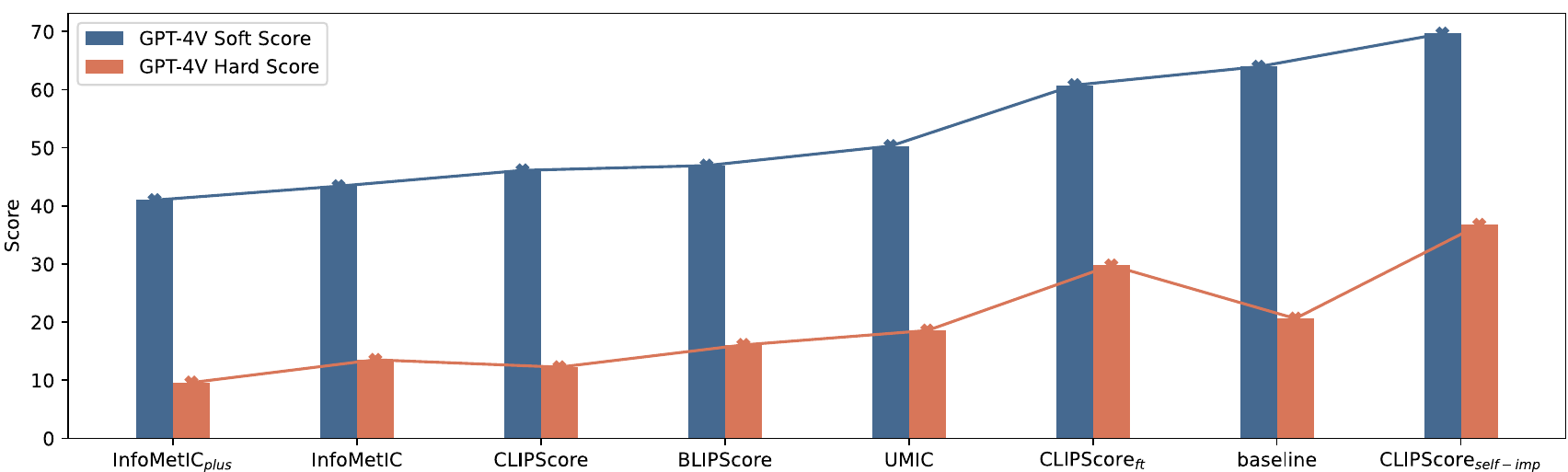}
    \caption{The results of GPT-4V evaluation.}
    \label{fig:gpt4v_score}
\end{figure*}

\subsection{GPT-4V Evaluation}
Recognizing the limitations inherent in traditional metrics like Cider \cite{DBLP:conf/cvpr/VedantamZP15}, which rely on n-gram analysis, we suggest that a low score under these criteria does not unequivocally denote poor sentence quality. The discrepancy may stem from significant variances in sentence structure rather than intrinsic quality. To avoid this issue, we have incorporated GPT-4V as an evaluative tool. Through carefully crafted prompts, we directed GPT-4V to assess the quality of sentences generated, assigning a score ranging from 0 to 1 based on multiple aspects, such as their coherence, relevance, and fluency.

Our evaluation encompasses sentences produced by eight distinct models: Baseline, UMIC, CLIPScore, CLIPScore${_\text{ft}}$, CLIPScore${_\text{self-imp}}$ (our proposed method), BLIPScore, InfoMetIC, and InfoMetIC$_{\text{plus}}$, utilizing a dataset of 5000 images from the test set. The outcomes of this assessment are delineated in \Cref{fig:gpt4v_score}. Here, the \texttt{GPT-4V Soft Score} refers to the nuanced scores ranging from 0 to 1, while the \texttt{GPT-4V Hard Score} is assigned based on a comparative metric, where the sentence receiving the highest score per image is awarded 1, and all others receive 0.

As depicted in \Cref{fig:gpt4v_score}, the Self-Improving method outperforms all compared metrics, achieving the highest ratings in both \texttt{GPT-4V Soft Score} and \texttt{GPT-4V Hard Score}. Specifically, our approach shows an improvement of 19.36 to 28.66 points on \texttt{GPT-4V Soft Score} and 18.18 to 27.14 points on \texttt{GPT-4V Hard Score} over conventional mainstream metrics. For a more equitable comparison, we also fine-tune a model, CLIPScore$_{\text{ft}}$, using Ground Truth data from the MSCOCO dataset. Relative to this fine-tuned model, our Self-Improving method registered an increase of 8.88 points on \texttt{GPT-4V Soft Score} and 6.98 points on \texttt{GPT-4V Hard Score}, underscoring its superior efficacy in curating negative samples.

Furthermore, when contrasted with the baseline model, our strategy demonstrated substantial gains—improving by 5.7 points on \texttt{GPT-4V Soft Score} and 16.12 points on \texttt{GPT-4V Hard Score}. This indicates that sentences generated through our method are of a higher quality than those produced by models solely relying on cross-entropy minimization with the Ground Truth. These findings affirm the substantial benefits of the Self-Improving method in generating more coherence.

\begin{table}
\centering
\small 
\resizebox{0.45\textwidth}{!}{
\begin{tabular}{lc}
\hline
\textbf{Metric} & \textbf{Score}\\
\hline
UMIC & 31.3 \\ 
BLIPScore & 58.4 \\ 
InfoMetIC & 31.8\\
InfoMetIC$_\text{plus}$ & 40.8 \\ \midrule
CLIPScore & 55.2 \\ 
CLIPScore$_\text{ft}$ & 56.3 \\
CLIPScore$_\text{self-imp}$ (ours) &  \textbf{96.6} \\

\hline
\end{tabular}
}
\caption{Resuls on Flaws Caption benchmark.}
\label{tb:flaws_caption_result}
\end{table}

\begin{table*}[h!]
\centering
\resizebox{0.99\textwidth}{!}{
\scriptsize
\begin{tabular}{c|m{.7\textwidth}}
\toprule
\textbf{Image} & \textbf{Generated Captions} \\
\midrule
% Example 1
\multirow{5}{*}{\begin{minipage}{.16\textwidth}\includegraphics[width=\linewidth]{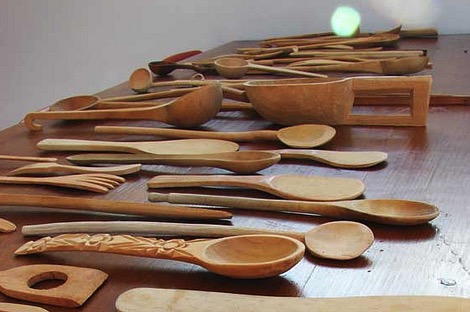}\end{minipage}} 
& \underline{\textbf{Baseline}} a wooden cutting board topped with wooden boards \\
& \\
& \underline{\textbf{CLIPScore$_\text{ft}$}} a wooden cutting board with wooden wooden wooden wooden wooden wooden wooden wooden wooden wooden wooden cutting board with wooden\\
& \\
& \underline{\textbf{CLIPScore$_\text{self-imp}$}} a wooden cutting board with some wooden utensils\\
\midrule

% Example 2
\multirow{5}{*}{\begin{minipage}{.15\textwidth}\includegraphics[width=\linewidth]{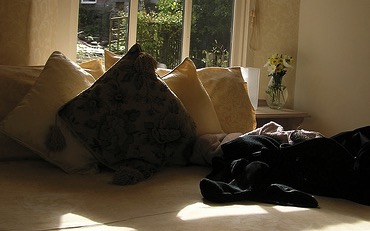}\end{minipage}} 
& \underline{\textbf{Baseline}} a living room with a couch and a window \\
& \\
& \underline{\textbf{CLIPScore$_\text{ft}$}} a bed sitting in a bedroom next to a window with a window view of a window \\
& \\
& \underline{\textbf{CLIPScore$_\text{self-imp}$}} a bed with pillows next to a window\\
\midrule

% Example 3
\multirow{5}{*}{\begin{minipage}{.16\textwidth}\includegraphics[width=\linewidth]{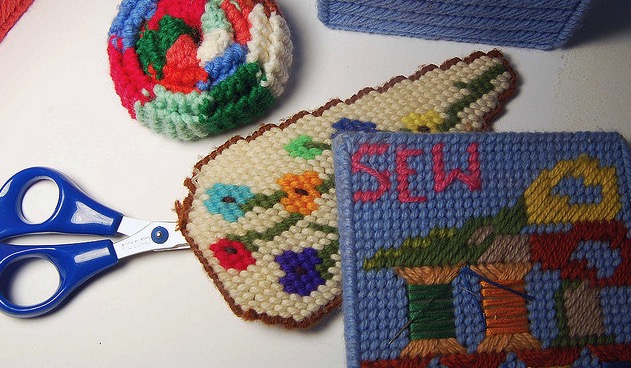}\end{minipage}} 
& \underline{\textbf{Baseline}} a blue pair of scissors next to a cake \\
& \\
& \underline{\textbf{CLIPScore$_\text{ft}$}} a blue pair of scissors next to a blue and blue cloth with a blue and\\
& \\
& \underline{\textbf{CLIPScore$_\text{self-imp}$}} a blue pair of scissors next to a blue cloth \\

\bottomrule
\end{tabular}
}
\caption{ Three Cases of generated sentences by Baseline, CLIPScore$_\text{ft}$, and CLIPScore$_\text{self-imp}$.}
\label{tb:case_study}
\end{table*}

\subsection{Flaws Caption benchmark Evaluation}
\label{sec:flaws_caption}
To assess the ability of metrics to accurately select appropriate captions for a given image amidst interference, we introduce a novel evaluation benchmark tailored for reference-free image captioning metrics, named Flaws Caption. 
This benchmark comprises 5,000 images, accompanied by a total of 50,000 descriptions. 
For each image, there are ten descriptions, evenly divided between five captions exhibiting flaws and five from the Ground Truth. 
The Flaws Caption benchmark requires evaluating the capability of a metric to discern one sentence that is most likely to be Ground Truth from among ten potential sentences.

We report the experiment results on Flaws Caption in \Cref{tb:flaws_caption_result}. Our observations reveal that the proposed method attained 96.6 points, markedly surpassing the performance of existing metrics. 
Notably, even though CLIPScore$_\text{ft}$ fine-tuned employing the MSCOCO dataset, it achieved a mere 56.3 points, only 1.1 percentage points higher than the original CLIPScore. These results underscore the superior efficacy of our Self-Improving approach in enhancing metric robustness for caption evaluation.

\subsection{Case Study}
In \Cref{tb:case_study}, we present a comparative analysis of generated sentences by our method against those produced by the Baseline and CLIPScore${_\text{ft}}$ models. 
It clearly demonstrates that descriptions generated by our method exhibit a higher degree of alignment with the corresponding images. 
For instance, an illustrative example from the bottom case in \Cref{tb:case_study} reveals that the Baseline model erroneously identifies an object as a  `cake'. 
While CLIPScore${_\text{ft}}$ corrects the object recognition error, the resulting sentences suffer from grammatical inconsistencies. 
In contrast, our CLIPScore${_\text{self-imp}}$ model not only accurately identifies the object as `cloth' but also produces grammatically fluent statements, underscoring the superiority of our approach in generating both accurate and linguistically coherent descriptions.

\section{Conclusion}
In this paper, we conduct a thorough examination of the field of reference-free image captioning metrics and find that prevalent reference-free captioning metrics have serious defects.
To address these issues, we introduce an innovative approach termed Self-Improving, which utilizes negative samples generated through reinforcement learning to improve the robustness of metrics. Furthermore, we establish a new evaluation benchmark named the Flaws Caption, towards mitigating the shortcomings of current metric evaluation.

\section{Limitation}
The current work has some limitations worth noting. First, limited by the action space, we use the MSCOCO dataset as the training set for captioning, constituting a vocabulary (action space) of about 10k words. Some works adopt the CC12M \citep{DBLP:conf/cvpr/ChangpinyoSDS21} as the training set, which has a larger vocabulary, presenting significant challenges for generating sentences through reinforcement learning \cite{DBLP:conf/icml/WangYMLBLMZZY22}. We will explore how to evaluate metrics in a larger vocabulary space in the future. Second, in the experiments, we chose the widely used vanilla Transformer as the backbone. Comparing models with performance diversity under the same metric might reveal more issues with the metrics. We plan to conduct this experiment in future work and compare the differences in outcomes.

\section{Acknowledgements}
This work was supported by NSFC No. 62176115.

\bibliography{custom}

\begin{thebibliography}{35}
\expandafter\ifx\csname natexlab\endcsname\relax\def\natexlab#1{#1}\fi

\bibitem[{Anderson et~al.(2016)Anderson, Fernando, Johnson, and Gould}]{DBLP:conf/eccv/AndersonFJG16}
Peter Anderson, Basura Fernando, Mark Johnson, and Stephen Gould. 2016.
\newblock \href {https://doi.org/10.1007/978-3-319-46454-1\_24} {{SPICE:} semantic propositional image caption evaluation}.
\newblock In \emph{Computer Vision - {ECCV} 2016 - 14th European Conference, Amsterdam, The Netherlands, October 11-14, 2016, Proceedings, Part {V}}, volume 9909 of \emph{Lecture Notes in Computer Science}, pages 382--398. Springer.

\bibitem[{Anderson et~al.(2018)Anderson, He, Buehler, Teney, Johnson, Gould, and Zhang}]{bottom-up}
Peter Anderson, Xiaodong He, Chris Buehler, Damien Teney, Mark Johnson, Stephen Gould, and Lei Zhang. 2018.
\newblock Bottom-up and top-down attention for image captioning and visual question answering.
\newblock In \emph{Proc. of CVPR}.

\bibitem[{Bai et~al.(2023)Bai, Bai, Yang, Wang, Tan, Wang, Lin, Zhou, and Zhou}]{DBLP:journals/corr/abs-2308-12966}
Jinze Bai, Shuai Bai, Shusheng Yang, Shijie Wang, Sinan Tan, Peng Wang, Junyang Lin, Chang Zhou, and Jingren Zhou. 2023.
\newblock \href {https://doi.org/10.48550/ARXIV.2308.12966} {Qwen-vl: {A} frontier large vision-language model with versatile abilities}.
\newblock \emph{CoRR}, abs/2308.12966.

\bibitem[{Banerjee and Lavie(2005)}]{DBLP:conf/acl/BanerjeeL05}
Satanjeev Banerjee and Alon Lavie. 2005.
\newblock \href {https://www.aclweb.org/anthology/W05-0909/} {{METEOR:} an automatic metric for {MT} evaluation with improved correlation with human judgments}.
\newblock In \emph{Proceedings of the Workshop on Intrinsic and Extrinsic Evaluation Measures for Machine Translation and/or Summarization@ACL 2005, Ann Arbor, Michigan, USA, June 29, 2005}, pages 65--72. Association for Computational Linguistics.

\bibitem[{Changpinyo et~al.(2021)Changpinyo, Sharma, Ding, and Soricut}]{DBLP:conf/cvpr/ChangpinyoSDS21}
Soravit Changpinyo, Piyush Sharma, Nan Ding, and Radu Soricut. 2021.
\newblock \href {https://doi.org/10.1109/CVPR46437.2021.00356} {Conceptual 12m: Pushing web-scale image-text pre-training to recognize long-tail visual concepts}.
\newblock In \emph{{IEEE} Conference on Computer Vision and Pattern Recognition, {CVPR} 2021, virtual, June 19-25, 2021}, pages 3558--3568. Computer Vision Foundation / {IEEE}.

\bibitem[{Chen et~al.(2023)Chen, Li, Dong, Zhang, He, Wang, Zhao, and Lin}]{DBLP:journals/corr/abs-2311-12793}
Lin Chen, Jinsong Li, Xiaoyi Dong, Pan Zhang, Conghui He, Jiaqi Wang, Feng Zhao, and Dahua Lin. 2023.
\newblock \href {https://doi.org/10.48550/ARXIV.2311.12793} {Sharegpt4v: Improving large multi-modal models with better captions}.
\newblock \emph{CoRR}, abs/2311.12793.

\bibitem[{Chen et~al.(2015)Chen, Fang, Lin, Vedantam, Gupta, Doll{\'a}r, and Zitnick}]{chen2015microsoft}
Xinlei Chen, Hao Fang, Tsung-Yi Lin, Ramakrishna Vedantam, Saurabh Gupta, Piotr Doll{\'a}r, and C~Lawrence Zitnick. 2015.
\newblock Microsoft coco captions: Data collection and evaluation server.
\newblock \emph{arXiv preprint arXiv:1504.00325}.

\bibitem[{Chen et~al.(2020)Chen, Li, Yu, Kholy, Ahmed, Gan, Cheng, and Liu}]{DBLP:conf/eccv/ChenLYK0G0020}
Yen{-}Chun Chen, Linjie Li, Licheng Yu, Ahmed~El Kholy, Faisal Ahmed, Zhe Gan, Yu~Cheng, and Jingjing Liu. 2020.
\newblock \href {https://doi.org/10.1007/978-3-030-58577-8\_7} {{UNITER:} universal image-text representation learning}.
\newblock In \emph{Computer Vision - {ECCV} 2020 - 16th European Conference, Glasgow, UK, August 23-28, 2020, Proceedings, Part {XXX}}, volume 12375 of \emph{Lecture Notes in Computer Science}, pages 104--120. Springer.

\bibitem[{Cheng et~al.(2023)Cheng, Song, Ma, Zhu, Zhu, and Zhang}]{DBLP:conf/mm/ChengSMZZZ23}
Kanzhi Cheng, Wenpo Song, Zheng Ma, Wenhao Zhu, Zixuan Zhu, and Jianbing Zhang. 2023.
\newblock \href {https://doi.org/10.1145/3581783.3611987} {Beyond generic: Enhancing image captioning with real-world knowledge using vision-language pre-training model}.
\newblock In \emph{Proceedings of the 31st {ACM} International Conference on Multimedia, {MM} 2023, Ottawa, ON, Canada, 29 October 2023- 3 November 2023}, pages 5038--5047. {ACM}.

\bibitem[{Cho et~al.(2022)Cho, Yoon, Kale, Dernoncourt, Bui, and Bansal}]{DBLP:conf/naacl/00010KDBB22}
Jaemin Cho, Seunghyun Yoon, Ajinkya Kale, Franck Dernoncourt, Trung Bui, and Mohit Bansal. 2022.
\newblock \href {https://doi.org/10.18653/V1/2022.FINDINGS-NAACL.39} {Fine-grained image captioning with {CLIP} reward}.
\newblock In \emph{Findings of the Association for Computational Linguistics: {NAACL} 2022, Seattle, WA, United States, July 10-15, 2022}, pages 517--527. Association for Computational Linguistics.

\bibitem[{Devlin et~al.(2019)Devlin, Chang, Lee, and Toutanova}]{DBLP:conf/naacl/DevlinCLT19}
Jacob Devlin, Ming{-}Wei Chang, Kenton Lee, and Kristina Toutanova. 2019.
\newblock \href {https://doi.org/10.18653/V1/N19-1423} {{BERT:} pre-training of deep bidirectional transformers for language understanding}.
\newblock In \emph{Proceedings of the 2019 Conference of the North American Chapter of the Association for Computational Linguistics: Human Language Technologies, {NAACL-HLT} 2019, Minneapolis, MN, USA, June 2-7, 2019, Volume 1 (Long and Short Papers)}, pages 4171--4186. Association for Computational Linguistics.

\bibitem[{Hessel et~al.(2021)Hessel, Holtzman, Forbes, Bras, and Choi}]{DBLP:journals/corr/abs-2104-08718}
Jack Hessel, Ari Holtzman, Maxwell Forbes, Ronan~Le Bras, and Yejin Choi. 2021.
\newblock \href {http://arxiv.org/abs/2104.08718} {Clipscore: {A} reference-free evaluation metric for image captioning}.
\newblock \emph{CoRR}, abs/2104.08718.

\bibitem[{Hodosh et~al.(2015)Hodosh, Young, and Hockenmaier}]{DBLP:conf/ijcai/HodoshYH15}
Micah Hodosh, Peter Young, and Julia Hockenmaier. 2015.
\newblock \href {http://ijcai.org/Abstract/15/593} {Framing image description as a ranking task: Data, models and evaluation metrics (extended abstract)}.
\newblock In \emph{Proceedings of the Twenty-Fourth International Joint Conference on Artificial Intelligence, {IJCAI} 2015, Buenos Aires, Argentina, July 25-31, 2015}, pages 4188--4192. {AAAI} Press.

\bibitem[{Hu et~al.(2023)Hu, Chen, Zhang, and Jin}]{DBLP:conf/acl/HuCZJ23}
Anwen Hu, Shizhe Chen, Liang Zhang, and Qin Jin. 2023.
\newblock \href {https://doi.org/10.18653/V1/2023.ACL-LONG.178} {Infometic: An informative metric for reference-free image caption evaluation}.
\newblock In \emph{Proceedings of the 61st Annual Meeting of the Association for Computational Linguistics (Volume 1: Long Papers), {ACL} 2023, Toronto, Canada, July 9-14, 2023}, pages 3171--3185. Association for Computational Linguistics.

\bibitem[{Jiang et~al.(2019)Jiang, Huang, Zhang, Wang, Zhang, Gan, Diesner, and Gao}]{DBLP:conf/emnlp/JiangHZWZGDG19}
Ming Jiang, Qiuyuan Huang, Lei Zhang, Xin Wang, Pengchuan Zhang, Zhe Gan, Jana Diesner, and Jianfeng Gao. 2019.
\newblock \href {https://doi.org/10.18653/V1/D19-1220} {Tiger: Text-to-image grounding for image caption evaluation}.
\newblock In \emph{Proceedings of the 2019 Conference on Empirical Methods in Natural Language Processing and the 9th International Joint Conference on Natural Language Processing, {EMNLP-IJCNLP} 2019, Hong Kong, China, November 3-7, 2019}, pages 2141--2152. Association for Computational Linguistics.

\bibitem[{Karpathy and Fei-Fei(2015)}]{karpathy2015deep}
Andrej Karpathy and Li~Fei-Fei. 2015.
\newblock Deep visual-semantic alignments for generating image descriptions.
\newblock In \emph{Proc. of CVPR}.

\bibitem[{Krishna et~al.(2017)Krishna, Zhu, Groth, Johnson, Hata, Kravitz, Chen, Kalantidis, Li, Shamma, Bernstein, and Fei{-}Fei}]{vg}
Ranjay Krishna, Yuke Zhu, Oliver Groth, Justin Johnson, Kenji Hata, Joshua Kravitz, Stephanie Chen, Yannis Kalantidis, Li{-}Jia Li, David~A. Shamma, Michael~S. Bernstein, and Li~Fei{-}Fei. 2017.
\newblock Visual genome: Connecting language and vision using crowdsourced dense image annotations.
\newblock \emph{Int. J. Comput. Vis.}

\bibitem[{Lee et~al.(2021)Lee, Yoon, Dernoncourt, Bui, and Jung}]{DBLP:conf/acl/Lee0DBJ20}
Hwanhee Lee, Seunghyun Yoon, Franck Dernoncourt, Trung Bui, and Kyomin Jung. 2021.
\newblock \href {https://doi.org/10.18653/V1/2021.ACL-SHORT.29} {{UMIC:} an unreferenced metric for image captioning via contrastive learning}.
\newblock In \emph{Proceedings of the 59th Annual Meeting of the Association for Computational Linguistics and the 11th International Joint Conference on Natural Language Processing, {ACL/IJCNLP} 2021, (Volume 2: Short Papers), Virtual Event, August 1-6, 2021}, pages 220--226. Association for Computational Linguistics.

\bibitem[{Li et~al.(2022)Li, Li, Xiong, and Hoi}]{DBLP:conf/icml/0001LXH22}
Junnan Li, Dongxu Li, Caiming Xiong, and Steven C.~H. Hoi. 2022.
\newblock \href {https://proceedings.mlr.press/v162/li22n.html} {{BLIP:} bootstrapping language-image pre-training for unified vision-language understanding and generation}.
\newblock In \emph{International Conference on Machine Learning, {ICML} 2022, 17-23 July 2022, Baltimore, Maryland, {USA}}, volume 162 of \emph{Proceedings of Machine Learning Research}, pages 12888--12900. {PMLR}.

\bibitem[{Li et~al.(2019)Li, Zhang, Li, Li, and Fu}]{DBLP:conf/iccv/LiZLLF19}
Kunpeng Li, Yulun Zhang, Kai Li, Yuanyuan Li, and Yun Fu. 2019.
\newblock \href {https://doi.org/10.1109/ICCV.2019.00475} {Visual semantic reasoning for image-text matching}.
\newblock In \emph{2019 {IEEE/CVF} International Conference on Computer Vision, {ICCV} 2019, Seoul, Korea (South), October 27 - November 2, 2019}, pages 4653--4661. {IEEE}.

\bibitem[{Lin(2004)}]{lin2004rouge}
Chin-Yew Lin. 2004.
\newblock Rouge: A package for automatic evaluation of summaries.
\newblock In \emph{Text summarization branches out}, pages 74--81.

\bibitem[{Liu et~al.(2023)Liu, Li, Wu, and Lee}]{DBLP:journals/corr/abs-2304-08485}
Haotian Liu, Chunyuan Li, Qingyang Wu, and Yong~Jae Lee. 2023.
\newblock \href {https://doi.org/10.48550/ARXIV.2304.08485} {Visual instruction tuning}.
\newblock \emph{CoRR}, abs/2304.08485.

\bibitem[{Lu et~al.(2019)Lu, Batra, Parikh, and Lee}]{DBLP:conf/nips/LuBPL19}
Jiasen Lu, Dhruv Batra, Devi Parikh, and Stefan Lee. 2019.
\newblock \href {https://proceedings.neurips.cc/paper/2019/hash/c74d97b01eae257e44aa9d5bade97baf-Abstract.html} {Vilbert: Pretraining task-agnostic visiolinguistic representations for vision-and-language tasks}.
\newblock In \emph{Advances in Neural Information Processing Systems 32: Annual Conference on Neural Information Processing Systems 2019, NeurIPS 2019, December 8-14, 2019, Vancouver, BC, Canada}, pages 13--23.

\bibitem[{Lu et~al.(2023)Lu, Yang, Li, Wang, and Wang}]{DBLP:journals/corr/abs-2305-11116}
Yujie Lu, Xianjun Yang, Xiujun Li, Xin~Eric Wang, and William~Yang Wang. 2023.
\newblock \href {https://doi.org/10.48550/ARXIV.2305.11116} {Llmscore: Unveiling the power of large language models in text-to-image synthesis evaluation}.
\newblock \emph{CoRR}, abs/2305.11116.

\bibitem[{Ma et~al.(2023)Ma, Wang, Huang, Zhu, and Zhang}]{DBLP:conf/nlpcc/MaWHZZ23}
Zheng Ma, Changxin Wang, Bo~Huang, Zixuan Zhu, and Jianbing Zhang. 2023.
\newblock \href {https://doi.org/10.1007/978-3-031-44693-1\_37} {Bounding and filling: {A} fast and flexible framework for image captioning}.
\newblock In \emph{Natural Language Processing and Chinese Computing - 12th National {CCF} Conference, {NLPCC} 2023, Foshan, China, October 12-15, 2023, Proceedings, Part {I}}, volume 14302 of \emph{Lecture Notes in Computer Science}, pages 469--481. Springer.

\bibitem[{Ma et~al.(2022)Ma, Zong, Pan, Zhang, Huang, Dai, and Chen}]{ma-etal-2022-probing}
Zheng Ma, Shi Zong, Mianzhi Pan, Jianbing Zhang, Shujian Huang, Xinyu Dai, and Jiajun Chen. 2022.
\newblock \href {https://doi.org/10.18653/v1/2022.findings-emnlp.421} {Probing cross-modal semantics alignment capability from the textual perspective}.
\newblock In \emph{Findings of the Association for Computational Linguistics: EMNLP 2022}, pages 5739--5749, Abu Dhabi, United Arab Emirates. Association for Computational Linguistics.

\bibitem[{Papineni et~al.(2002)Papineni, Roukos, Ward, and Zhu}]{DBLP:conf/acl/PapineniRWZ02}
Kishore Papineni, Salim Roukos, Todd Ward, and Wei{-}Jing Zhu. 2002.
\newblock Bleu: a method for automatic evaluation of machine translation.
\newblock In \emph{Proceedings of the 40th Annual Meeting of the Association for Computational Linguistics, July 6-12, 2002, Philadelphia, PA, {USA}}, pages 311--318. {ACL}.

\bibitem[{Radford et~al.(2021)Radford, Kim, Hallacy, Ramesh, Goh, Agarwal, Sastry, Askell, Mishkin, Clark, Krueger, and Sutskever}]{DBLP:conf/icml/RadfordKHRGASAM21}
Alec Radford, Jong~Wook Kim, Chris Hallacy, Aditya Ramesh, Gabriel Goh, Sandhini Agarwal, Girish Sastry, Amanda Askell, Pamela Mishkin, Jack Clark, Gretchen Krueger, and Ilya Sutskever. 2021.
\newblock \href {http://proceedings.mlr.press/v139/radford21a.html} {Learning transferable visual models from natural language supervision}.
\newblock In \emph{Proceedings of the 38th International Conference on Machine Learning, {ICML} 2021, 18-24 July 2021, Virtual Event}, volume 139 of \emph{Proceedings of Machine Learning Research}, pages 8748--8763. {PMLR}.

\bibitem[{Rennie et~al.(2017)Rennie, Marcheret, Mroueh, Ross, and Goel}]{DBLP:conf/cvpr/RennieMMRG17}
Steven~J. Rennie, Etienne Marcheret, Youssef Mroueh, Jerret Ross, and Vaibhava Goel. 2017.
\newblock \href {https://doi.org/10.1109/CVPR.2017.131} {Self-critical sequence training for image captioning}.
\newblock In \emph{2017 {IEEE} Conference on Computer Vision and Pattern Recognition, {CVPR} 2017, Honolulu, HI, USA, July 21-26, 2017}, pages 1179--1195. {IEEE} Computer Society.

\bibitem[{Shi et~al.(2019)Shi, Ji, Lu, Niu, and Duan}]{DBLP:conf/ijcai/ShiJLND19}
Botian Shi, Lei Ji, Pan Lu, Zhendong Niu, and Nan Duan. 2019.
\newblock \href {https://doi.org/10.24963/IJCAI.2019/720} {Knowledge aware semantic concept expansion for image-text matching}.
\newblock In \emph{Proceedings of the Twenty-Eighth International Joint Conference on Artificial Intelligence, {IJCAI} 2019, Macao, China, August 10-16, 2019}, pages 5182--5189. ijcai.org.

\bibitem[{Varpio et~al.(2017)Varpio, Ajjawi, Monrouxe, O'Brien, and Rees}]{varpio2017shedding}
Lara Varpio, Rola Ajjawi, Lynn~V Monrouxe, Bridget~C O'Brien, and Charlotte~E Rees. 2017.
\newblock Shedding the cobra effect: problematising thematic emergence, triangulation, saturation and member checking.
\newblock \emph{Medical education}, 51(1):40--50.

\bibitem[{Vaswani et~al.(2017)Vaswani, Shazeer, Parmar, Uszkoreit, Jones, Gomez, Kaiser, and Polosukhin}]{transformer}
Ashish Vaswani, Noam Shazeer, Niki Parmar, Jakob Uszkoreit, Llion Jones, Aidan~N Gomez, {\L}ukasz Kaiser, and Illia Polosukhin. 2017.
\newblock Attention is all you need.
\newblock \emph{Proc. of NeurIPS}.

\bibitem[{Vedantam et~al.(2015)Vedantam, Zitnick, and Parikh}]{DBLP:conf/cvpr/VedantamZP15}
Ramakrishna Vedantam, C.~Lawrence Zitnick, and Devi Parikh. 2015.
\newblock Cider: Consensus-based image description evaluation.
\newblock In \emph{{IEEE} Conference on Computer Vision and Pattern Recognition, {CVPR} 2015, Boston, MA, USA, June 7-12, 2015}, pages 4566--4575. {IEEE} Computer Society.

\bibitem[{Wang et~al.(2022)Wang, Yang, Men, Lin, Bai, Li, Ma, Zhou, Zhou, and Yang}]{DBLP:conf/icml/WangYMLBLMZZY22}
Peng Wang, An~Yang, Rui Men, Junyang Lin, Shuai Bai, Zhikang Li, Jianxin Ma, Chang Zhou, Jingren Zhou, and Hongxia Yang. 2022.
\newblock \href {https://proceedings.mlr.press/v162/wang22al.html} {{OFA:} unifying architectures, tasks, and modalities through a simple sequence-to-sequence learning framework}.
\newblock In \emph{International Conference on Machine Learning, {ICML} 2022, 17-23 July 2022, Baltimore, Maryland, {USA}}, volume 162 of \emph{Proceedings of Machine Learning Research}, pages 23318--23340. {PMLR}.

\bibitem[{Yuksekgonul et~al.(2022)Yuksekgonul, Bianchi, Kalluri, Jurafsky, and Zou}]{yuksekgonul2022and}
Mert Yuksekgonul, Federico Bianchi, Pratyusha Kalluri, Dan Jurafsky, and James Zou. 2022.
\newblock When and why vision-language models behave like bags-of-words, and what to do about it?
\newblock In \emph{The Eleventh International Conference on Learning Representations}.

\end{thebibliography}
\appendix

\end{document}